\documentclass[preprint,12pt]{elsarticle}

\usepackage{amssymb}
\usepackage{amsmath}
\usepackage{hyperref}
\usepackage{colortbl}
\definecolor{Gray}{gray}{0.93}
\usepackage{multirow}
\journal{Information Fusion}

\begin{document}

\begin{frontmatter}



\title{CoreNet: Conflict Resolution Network for Point-Pixel Misalignment and Sub-Task Suppression of 3D LiDAR-Camera Object Detection\tnoteref{1}}

\tnotetext[1]{Yiheng Li and Yang Yang contribute equally to this work.}

\author[label1,label2]{Yiheng Li}
\ead{liyiheng2024@ia.ac.cn}
\author[label1,label2]{Yang Yang\corref{cor1}} 
\ead{yang.yang@nlpr.ia.ac.cn}
\author[label1,label2,label3]{Zhen Lei}
\ead{zhen.lei@ia.ac.cn}
\affiliation[label1]{organization={ State Key Laboratory of Multimodal Artificial Intelligence Systems (MAIS), Institute of Automation},
            addressline={Chinese Academy of Sciences},
            city={Beijing},
            postcode={100190},
            country={China}}
\affiliation[label2]{organization={School of Artificial Intelligence},
            addressline={University of Chinese Academy of Sciences},
            city={Beijing},
            postcode={100049},
            country={China}}
\affiliation[label3]{organization={Centre for Artificial Intelligence and Robotics, Hong Kong Institute of Science \&
Innovation},
            addressline={Chinese Academy of Sciences},
            city={Hongkong},
            postcode={999077},
            country={China}}
            
\cortext[cor1]{Corresponding author}

% \affiliation[label2]{organization={},

% \author{Yiheng Li} %% Author name

% \affiliation{organization={},%Department and Organization

%% Abstract
\begin{abstract}
Fusing multi-modality inputs from different sensors is an effective way to improve the performance of 3D object detection. However, current methods overlook two important conflicts: point-pixel misalignment and sub-task suppression. The former means a pixel feature from the opaque object is projected to multiple point features of the same ray in the world space, and the latter means the classification prediction and bounding box regression may cause mutual suppression. In this paper, we propose a novel method named \textbf{Co}nflict \textbf{Re}solution \textbf{Net}work (CoreNet) to address the aforementioned issues. Specifically, we first propose a dual-stream transformation module to tackle point-pixel misalignment. It consists of ray-based and point-based 2D-to-BEV transformations. Both of them achieve approximately unique mapping from the image space to the world space. Moreover, we introduce a task-specific predictor to tackle sub-task suppression. It uses the dual-branch structure which adopts class-specific query and Bbox-specific query to corresponding sub-tasks. Each task-specific query is constructed of task-specific feature and general feature, which allows the heads to adaptively select information of interest based on different sub-tasks. Experiments on the large-scale nuScenes dataset demonstrate the superiority of our proposed CoreNet, by achieving 75.6\% NDS and 73.3\% mAP on the nuScenes test set without test-time augmentation and model ensemble techniques. The ample ablation study also demonstrates the effectiveness of each component. The code is released on \url{https://github.com/liyih/CoreNet}.

\end{abstract}


% \begin{highlights}
% \item We propose a novel multi-modality 3D object detection framework named CoreNet, which focuses on tackling the conflicts in the fusion and prediction stages.
% \item We design dual-stream transformation module to solve the point-pixel misalignment and a task-specific predictor to solve the sub-task suppression.
% \item We confirm the efficacy of the proposed CoreNet framework for LiDAR-camera 3D object detection by achieving the superiority performance on large-scale nuScenes dataset.
% \end{highlights}

%% Keywords

\end{frontmatter}


\section{Introduction}
\label{sec1}
%% Labels are used to cross-reference an item using \ref command.

\begin{figure*}[h!]%% placement specifier
\centering
\includegraphics[scale=0.5]{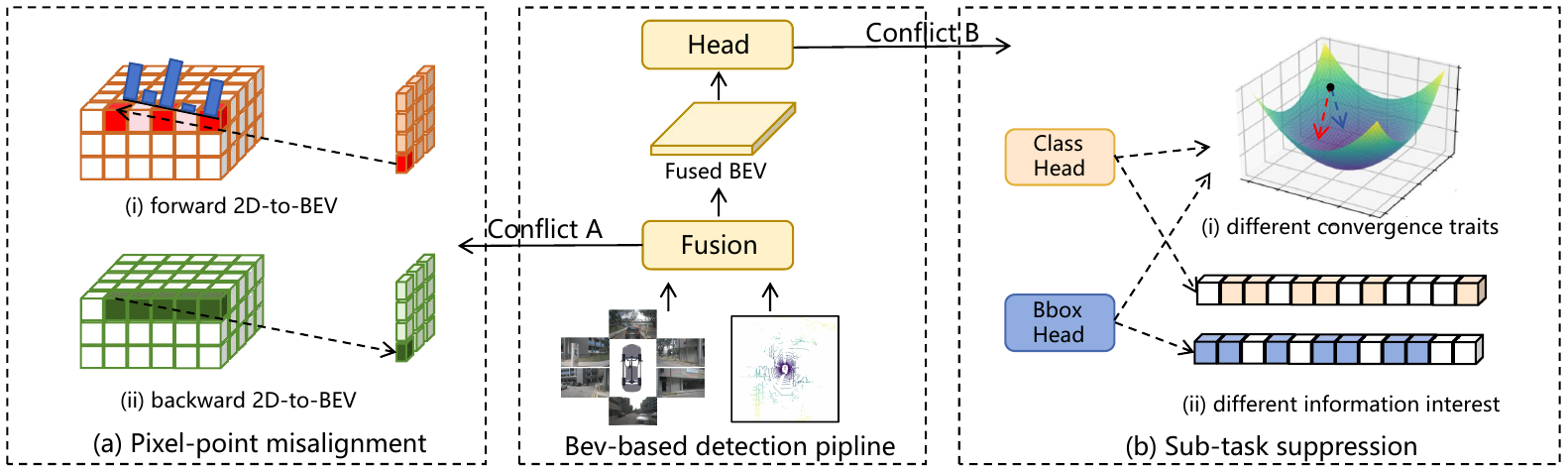}
\caption{Display of two conflicts for 3D LiDAR-camera object detection. The schematic diagram of point-pixel alignment is provided on the left, while the schematic diagram of sub-task suppression is shown on the right.}\label{fig1}
\end{figure*}

% 3D object detection is an important vision task, which plays a critical role in autonomous driving and robotics \cite{shi2019pointrcnn,yan2018second,yin2021center}. High-quality object detection results can be used as reliable observations for downstream tasks such as object tracking and path planning \cite{mao20233d}. The purpose of 3D object detection is to predict the locations, sizes, categories, orientations, and velocities of 3D objects around the ego vehicle. In recent years, multi-modality 3D object detection has received more and more attention as the requirements of robustness and high performance \cite{bai2022transfusion,liu2023bevfusion,cai2023objectfusion}. Camera and LiDAR are the two most common sensors. Camera can provide rich semantic information, while LiDAR can provide precise structure information. Compared to traditional single-modality 3D detection methods, multi-modality methods can often achieve superior and more robust results, mainly because of the rich information representation capability of multiple modalities and the ability of modalities to complement each other \cite{mao20233d}.

3D object detection is to predict the locations, sizes, categories, orientations, and velocities of 3D objects around the ego vehicle, which plays a critical role in autonomous driving and robotics \cite{huang2023multi, fernandes2021point, zhang2023dual}. High-quality object detection results can be used as reliable observations for downstream tasks such as object tracking and path planning \cite{wang2024deep, mao20233d}. In recent years, multi-modality 3D object detection has received more and more attention due to its robustness and high performance \cite{wu2024fusion,hao2024coarse,jiang2024mshp3d,yang2024v2x}. Camera and LiDAR are the two most common sensors. The former can provide rich semantic information, the latter can capture the precise structure of the objects. Compared to traditional single-modality 3D object detection, the application of multiple sensors can often achieve superior and more robust results. This is because multiple modalities have rich information representation capability and complement each other \cite{mao20233d, song2024robofusion, xu2024multi,jiang2024sparseinteraction, li2025rctrans}.

As the inputs from different modalities have heterogeneous data representations, most existing methods \cite{liu2023bevfusion, yin2024fusion} first project different modalities into a unified space e.g. the bird’s eye view (BEV), and then perform the fusion process. However, they often overlook two important conflicts, i.e., point-pixel misalignment and sub-task suppression (as shown in Fig. \ref{fig1}). 
Point-pixel misalignment appears in the 2D-to-BEV transformation stage. The current 2D-to-BEV transformation is divided into forward \cite{philion2020lift} and backward \cite{li2022bevformer} methods. Both of them project a pixel feature from an opaque object to multiple points of the pixel-object ray which connects the object and the pixel feature in the camera system. In reality, each pixel feature of opacity objects should be uniquely projected or mostly concentrated to a single point in the pixel-object ray.
The forward methods first predict the distribution probability among the uniformly sampled points of the ray from a pixel. Then, they scatter the pixel feature to the sampled points based on the distribution probability. Due to the fact that the probability is not a one-hot distribution, a pixel feature will be projected to multiple points of the same pixel-object ray. For backward methods, each point in the world space will be aligned with the corresponding pixel via the camera parameter. In this way, a pixel feature will be scattered to all the sampled points of the same pixel-object ray.
Sub-tasks suppression appears in the prediction stage. The task of 3D object detection can be divided into two sub-tasks: classification prediction and bounding box (Bbox) regression. 
They are supervised with different kinds of functions, e.g., classification often relies on cross entropy loss and bounding box regression uses L1 Loss.
These two sub-tasks may have different convergence directions and speeds, causing mutual suppression between sub-tasks during the joint training. In addition, different information of interest between sub-tasks will also give rise to the sub-tasks suppression \cite{zhang2023decoupled}.  Thus, how to address the problems of point-pixel misalignment and sub-task suppression have always been the core concerns.

In this paper, we propose a novel LiDAR-camera \textbf{Co}nflict \textbf{Re}solution \textbf{Net}work (CoreNet) to tackle the mentioned conflicts. For one thing, we introduce dual-stream transformation to overcome the point-pixel misalignment. It consists of ray-based and point-based 2D-to-BEV transformations. For the ray-based branch, inspired by BEVDepth \cite{li2023bevdepth}, we adopt the one-hot distribution to supervise the scattered probability along each pixel-object ray. It will guarantee a pixel feature is mostly concentrated to a single point in the pixel-object ray. For the point-based branch, we adopt the point cloud from LiDAR as the guidance. LiDAR point clouds mainly distribute on the surface of the opaque objects, and the occlusion relationship between the captured objects is basically the same as that of multi-view cameras. Based on this characteristic, each ray emitted by LiDAR will only return a unique landing signal to the sensor (located on opaque obstacles or backgrounds). We divide the point clouds into several bins, and each point of the same bin is located in the same BEV grid. The points in the same bin are projected to the image space. Corresponding pixel features are considered as their features which are pooled to be the BEV grid feature. 
For another, we propose a task-specific predictor to tackle the sub-task suppression. Most 3D object detections typically use separate prediction heads to extract sub-task features from the same feature \cite{bai2022transfusion, liu2023bevfusion, huang2021bevdet}. The concerns of different heads are different during feature extraction, which may reduce the sub-task suppression in the supervision and prediction. However, such a strategy of eliminating the suppression between sub-tasks is implicit and uncontrollable. 
Some methods \cite{hou2024query, deng2022vista} adopt the decoupling strategy which uses different features for classification and regression learning. Suppression mainly comes from different supervised signals, but there may be still some correlations between different sub-tasks. For example, the size and velocity of an object can reflect its type to an extent. 
Although the influence of supervised signals can be reduced by using fully separate features as previous decoupling manners, this may lead to difficulty in learning the relationship between classification and regression. In contrast, we extract general and specific features. General features are able to capture the correlations of sub-tasks and are used for both classification and regression, while specific features contain the information that is preferred by a certain sub-task and is used for a specific prediction head.
Specifically, We first explicitly construct features with different task preferences, and task-specific supervision is introduced to make this process more controllable. Then, the general feature is extracted for both prediction heads. After that, a task-specific fuser combines the task-specific and general features to obtain the task-specific query, which adaptively selects information of interest. Finally, the task-specific query is inputted to the corresponding sub-task head. 

Our proposed CoreNet achieves superior performance on the large-scale nuScenes \cite{caesar2020nuscenes} dataset by achieving 75.6\% NDS and 73.3\% mAP on the nuScenes test set.
It also beats the latest state-of-the-art approaches (i.e., IS-Fusion \cite{yin2024fusion} and DAL \cite{huang2023detecting}) on both nuScenes test and val set.
Ablation experiments demonstrate the effectiveness of our proposed dual-stream transformation and task-specific predictor.
% This phenomenon fully proves the correctness of the standpoint of this paper: using  different modality preference features for each sub-task can mitigate the mutual inhibition and achieve improved performance. 
The main contributions are as follows:
(1) We propose a novel multi-modality 3D object detection framework named CoreNet, which focuses on tackling the conflicts in the fusion and prediction stages.
(2) We design the dual-stream transformation and task-specific predictor to solve the point-pixel misalignment and to solve the sub-task suppression, respectively.
(3) We confirm the efficacy of the proposed CoreNet framework for LiDAR-camera 3D object detection by achieving superior performance on large-scale nuScenes dataset.

\section{Related Work}
\subsection{Camera-Based 3D object detection}

Camera-based 3D object detection is a fast and low economic cost solution in autonomous driving. Typically, it can be categorized into monocular and multi-view methods. For monocular detection, researchers directly attach prediction heads to 2D image backbone \cite{yin2021center,wang2021fcos3d}. Depth estimation sometimes accompanies these methods \cite{xu2023mononerd}. For multi-view detection which has higher accuracy and stronger perception capability, it can be divided into BEV-based \cite{liu2023sparsebev,li2023bevdepth,li2022bevformer,huang2021bevdet} and DETR-based methods \cite{liu2022petr,wang2023exploring, doll2022spatialdetr}. The former first extracts bird's-eye-view (BEV) features based on image features through view transformation. View transformation can be categorized into forward \cite{philion2020lift} and backward \cite{li2022bevformer} transformations. Then, 3D objects are predicted based on the BEV features. The latter firstly interacts with the object queries with the image tokens through attention mechanism\cite{vaswani2017attention,carion2020end}, and then predicts the 3D objects based on the transformer outputs. Some methods combine both forward and backward transformations. For example, FB-BEV \cite{li2023fb} proposes a Depth-Aware
Backward Projection to enhance the depth consistency between forward and backward transformations, thus obtaining high-quality BEV representations. 

\subsection{LiDAR-Based 3D object detection}

LiDAR-based 3D object detection can predict reliable results from the captured 3D structural information. Current LiDAR-based 3D object detection can be divided into point-based \cite{li2021lidar,qi2017pointnet,qi2017pointnet++,shi2019pointrcnn,qi2018frustum} and transformation-based  \cite{chen2022focal,chen2023largekernel3d,zhou2018voxelnet,yan2018second,li2023pillarnext,lang2019pointpillars,yin2021center,fan2021rangedet} methods.
Point-based methods directly extract features from raw point clouds and predict objects based on point-level features. The main challenge of these methods is the irregular format of the point cloud. Researchers construct many network structures, such as PointNet \cite{qi2017pointnet}, to solve this problem. Transformation-based methods usually transform the original point clouds into a regular format, such as voxels \cite{chen2022focal,chen2023largekernel3d,zhou2018voxelnet,yan2018second} and pillars \cite{li2023pillarnext,lang2019pointpillars,yin2021center}. After that, the 3D bounding box is predicted based on the transformed inputs.
The main advantage of these methods is that they can use flexible and mature standard convolutional networks \cite{lang2019pointpillars} or sparse 3D convolutional networks \cite{yan2018second} to process features. However, the transformation process of data format may lead to information loss. 

\subsection{LiDAR-Camera 3D object detection}

% Multi-modality 3D object detection is a robust and high-performance solution. 
Mutli-modality 3D object detection uses camera and LiDAR sensors to capture texture and geometry information, respectively. Thus, it achieves more robust and higher performance than uni-modality 3D object detection, i.e., camera-based or LiDAR-based approach.
How to efficiently fuse heterogeneous data is the key point for multi-modality 3D object detection \cite{bi2024dyfusion, song2024voxelnextfusion,xu2021fusionpainting}. 
Early works often use the LiDAR branch as the leading role for prediction, with camera inputs only serving as a way to supplement information. Pointpainting \cite{vora2020pointpainting} uses the segmentation masks of the images as a new feature dimension for the point cloud. MVP \cite{yin2021multimodal} uses the images to generate virtual points to complement the LiDAR point clouds, which can give objects a more complete contour. Transfusion \cite{bai2022transfusion} first predicts the results by the LiDAR branch and then refines the outputs based on the image inputs. The effectiveness of these methods is overly dependent on the LiDAR branch. When there are problems with the LiDAR branch, this type of method will not be able to achieve correct results. In view of this, the subsequent works equally emphasize both modalities.

Recently, researchers have begun to focus on finding a feature-level fusion, which can fully extract and integrate information from different modalities.
MVX-Net \cite{sindagi2019mvx} proposes pointfusion and voxelfusion strategies to make better interactions. The former is an early-stage fusion, while the latter is a late-stage fusion. 
% 3D-CVF \cite{yoo20203d} proposes a gated camera-liDAR fusion method and performs multi-modality fusion in both the feature and the ROI stage. 
To improve the robustness of feature fusion in difficult scenarios, Autoalignv2 \cite{chen2022deformable} proposes cross-modality attention with soft correspondence. 
DeepInteraction \cite{yang2022deepinteraction} captures and maintains unique information from each modality. The unique information is fully exploited and integrated during object detection. 
% UVTR \cite{li2022unifying} uses 3D voxel space as the unified representation space during fusion. 
BEVfusion \cite{liu2023bevfusion,liang2022bevfusion} projects the inputs of two modalities into BEV space to facilitate fusion. 
% SparseFusion \cite{xie2023sparsefusion} first predicts the candidates of each modality and then fuses these sparse candidates by self-attention. 
SparseFusion \cite{xie2023sparsefusion} firstly extracts candidate features in each modality and then fuses these candidates by a standard self-attention module.
CMT \cite{yan2023cross} projects different modalities into tokens and fuses them by a transformer decoder. UniTR \cite{wang2023unitr} uses a unified transformer encoder to fuse multi-modality inputs simultaneously. IS-Fusion \cite{yin2024fusion} conducts hierarchical scene fusion and instance-guided fusion to capture contextual information of different scales. GraphBEV \cite{song2024graphbev} utilizes the deformable module to tackle the misalignment caused by inaccurate calibration relationship. DAL \cite{huang2023detecting} imitates the labeling process to conduct feature fusion and object prediction. The core of these methods is to find a unified space for feature fusion, including BEV, voxel, token, etc. Currently, compared to early LiDAR-camera based methods, this type of method shows the ability to solve difficult situations and achieves better performance.

Despite the significant progress of fusing architecture, there is still a gap in eliminating the pixel-point misalignment and the mutual suppression of different sub-tasks. For the first conflict, existing approaches always overlook it or use depth supervision \cite{li2023bevdepth}. For the second conflict, existing approaches always adopt decoupled or separate heads \cite{zhang2023decoupled, zhuang2023task} that are used in 2D or single-modality detection without hesitation. In contrast, we design two sophisticated modules to tackle these conflicts.

\subsection{Misalignment issue}
The misalignment of different modalities is an important concern in LiDAR-camera 3D object detection \cite{song2024robustness, wang2023multi}, which could be roughly divided into alignment offset and non-uniqueness alignment. The former happens due to the inaccurate camera calibration, making the correspondences between 2D and 3D have some offsets. To solve it, researchers tend to use approaches such as projection offsets and neighboring projections \cite{song2024robustness}.
Specifically, GraphAlign \cite{song2023graphalign} and GraphAlign++ \cite{song2023graphalign++}
conduct local graph modeling and project it onto the image plane. Autoalign \cite{chen2022autoalign} maps cross-modal relation with a learnable alignment map. AutoalignV2 \cite{chen2022deformable} adopts learnable sampling points to model the cross-modality relation. GraphBEV \cite{song2024graphbev} uses neighbor depth features and deformable module to tackle alignment offsets.
ContrastAlign \cite{song2024contrastalign} enhances the alignment of modalities via contrastive learning.
Unlike the above-mentioned methods, in this paper, we focus on non-uniqueness alignment, that is, for opaque objects, the features of one pixel are mapped to multiple points on its corresponding pixel ray. Some uni-modality detections propose a similar conception and call it depth ambiguity. For example, DFA3D \cite{li2023dfa3d} extends 2D image features into 3D feature maps and uses 3D deformable attention to aggregate them, which alleviates the depth ambiguity problem from the root. In contrast, we use a more intuitive dual-stream method. In each branch, pixels are mapped to approximately unique points in the 3D space. Thus, our method can reduce the incorrect correspondences between LiDAR point clouds and images.

\section{Method}

\begin{figure*}[t]%% placement specifier
\centering
\includegraphics[scale=0.45]{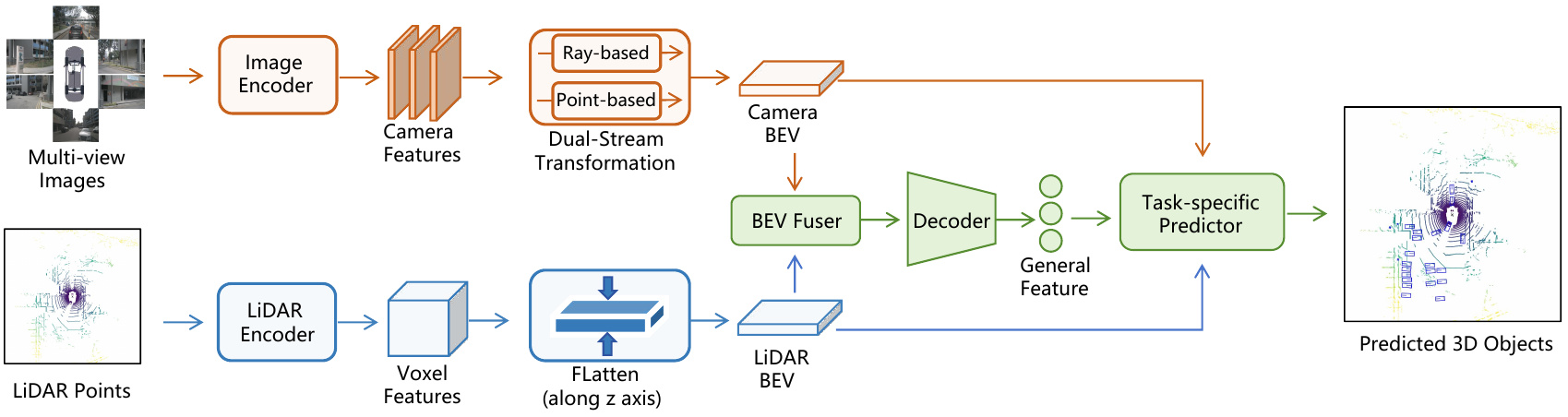}
\caption{The overall framework of our proposed CoreNet. The key components of our method are: (1) dual-stream transformation which tackles the point-pixel misalignment in 2D-to-BEV transformation, (2) task-specific predictor which tackles the sub-task suppression via explicitly constructing task-specific query.}\label{fig2}
\end{figure*}

The overall architecture of our proposed CoreNet is shown in Fig. \ref{fig2}. In this section, we first introduce the overall architecture of our proposed CoreNet in Sec. \ref{sec:3.1}. After that, we show the details of the dual-stream transformation module in Sec. \ref{sec:3.2}. Finally, we elaborate on the key components of the task-specific predictor in Sec. \ref{sec:3.3}.

\subsection{Overall Architecture}\label{sec:3.1}
We first project camera and LiDAR inputs into BEV features to facilitate subsequent steps. The reasons for selecting BEV features are as follows: (1) they are conducive to performing subsequent perception tasks, and (2) they allow for the effective use of mature 2D feature neural networks to deeply process multi-modality features.

For the LiDAR branch, we voxelize the point clouds ${P \in {R}^{N_p \times 5}}$ to obtain voxel feature ${V \in {R}^{X\times Y \times Z \times C_{v}}}$, where $N_p$ is the number of the points. Then, we use 3D sparse convolutional blocks \cite{yan2018second} as LiDAR encoder to obtain middle feature ${M \in {R}^{X\times Y \times Z \times C_{m}}}$. Finally, we compress the feature $M$ in Z-axis to obtain LiDAR BEV ${ B_l \in {R}^{X\times Y \times C_{l}}}$. For the camera branch, the multi-view image inputs ${I \in {R}^{N_i \times H\times W \times 3}}$ are encoded into multi-view features ${F \in {R}^{N_i \times H^{'}\times W^{'} \times C_f}}$ by the camera encoder, where $N_i$ is the number of images. 
After that, we obtain the camera BEV ${ B_c \in {R}^{X\times Y \times C_{c}}}$ based on the feature $F$ through dual-stream transformation.

Given the camera and LiDAR BEV, we input them into the BEV fuser to obtain the fused BEV. Following previous methods \cite{liu2023bevfusion, liang2022bevfusion}, the BEV fuser consists of the CNN-based encoder and SECONDFPN \cite{yan2018second} neck. Then, the fused BEV is inputted into the decoder head \cite{bai2022transfusion} to obtain general features of objects. Specifically, given the fused BEV $B_f$, a CNN-based heatmap encoder is first used to extract category-aware heatmap $H \in R^{X\times Y\times N}$ \cite{yin2021center,bai2022transfusion}, where $X\times Y$ is the size of BEV and $N$ is the number of categories. We select top $K$ local maximum elements as the object candidates, whose values are greater than or equal to their 8 nearest neighbors. 
Different from GraphBEV \cite{song2024graphbev} and GraphAlign \cite{song2023graphalign}, both of which use K nearest neighbors in the feature alignment stage and aim to alleviate the impact of inaccurate camera calibration, our method uses K nearest neighbors in the candidate selecting process to prevent the selected candidates from being too dense or clustered.
We denote the indexes of these candidates as $I_p=\{i_k\}_{k=1}^K$. At least, a transformer decoder is used to get the contextual information related to the candidates. The outputs of the transformer decoder are the general features $f_g$ which are universal and could be used in both sub-task heads.

Finally, the camera BEV $B_c$, LiDAR BEV $B_l$, and the general feature $f_g$ are inputted into the task-specific predictor to predict the final 3D objects.

\subsection{Dual-Stream Transformation}\label{sec:3.2}

\begin{figure*}[t]%% placement specifier
\centering
\includegraphics[scale=0.44]{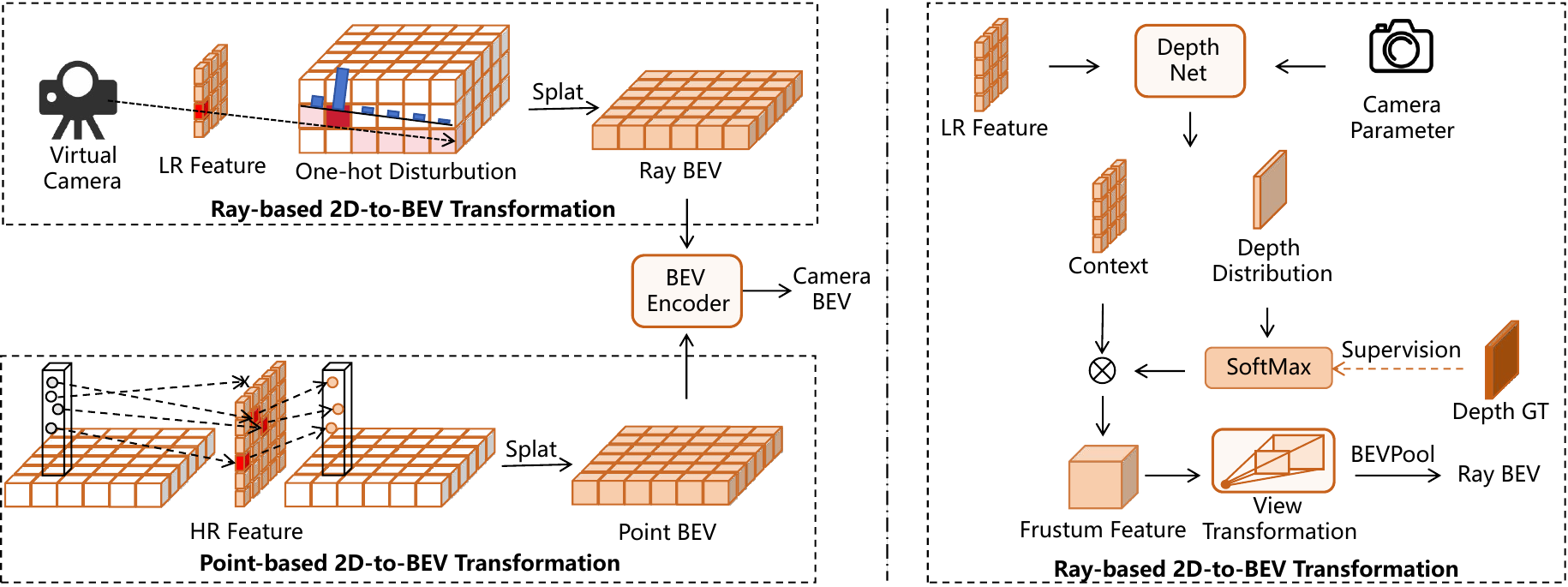}
\caption{The overall framework of our proposed dual-stream transformation.}\label{fig3}
\end{figure*}

The purpose of the dual-stream transformation is to make a pixel feature from the opaque object projected or approximately scattered into the single point of the same pixel-object ray in the world space. This way will better conform to the occlusion relationship between distant and nearby objects in the real world, thereby reducing the impact of redundant information on feature fusion. As shown in Fig. \ref{fig3}, dual-stream transformation consists of ray-based and point-based 2D-to-BEV transformations. Given the multi-view camera features $F$, we upsample them to the same shape as the image inputs via deconvolution block to obtain a high-resolution (HR) feature. Meanwhile, the channel of the feature map is reduced to decrease resource consumption. The HR feature is inputted to a point-based stream to make it create more precise connections between the point and the pixel, while the origin low-resolution (LR) feature $F$ is inputted to ray stream as too large feature sizes can consume a lot of computation time.
Each stream will output its corresponding BEV. The CNN-based BEV encoder will fuse the ray and point BEV and output the final camera BEV.

\subsubsection{Ray-Based 2D-to-BEV Transformation}
Given the LR feature, following BEVDepth \cite{li2023bevdepth}, we adopt depth net to encode the LR feature and camera parameter. The output of depth net is the context feature $F_t \in R^{N_i\times H^{'} \times W^{'} \times C_t }$ and the depth distribution $D_p \in R^{N_i\times H^{'} \times W^{'} \times D}$, where $D$ is the number of points along the depth axis. We suppose that the depth distribution of each pixel should be close to the one-hot distribution because of opacity and occlusion. In view of this, we generate the one-hot depth ground-truth $D_g$ based on the LiDAR point clouds and camera parameters. The depth loss can be calculated via Eq. 1.
\begin{equation}\label{label:1}
L_{depth} = \Phi (D_g, D_p),
\end{equation}
where $\Phi$ is the binary cross entropy loss. Based on $F_t$ and $D_p$, we obtain frustum feature ${T \in {R}^{N_i \times H\times W \times D \times C_{t}}}$. At last, ${T}$ is projected into world space, and the information in the Z-axis is compressed using BEVPool \cite{liu2023bevfusion} to get the final ray BEV.
\subsubsection{Point-Based 2D-to-BEV Transformation}
In the point-based stream, we use the LiDAR points which are located on the surface of the objects as the guidance. We suppose it can help the pixel features be projected to the correct position in the scene.
We first partition the scene into the BEV grids, each BEV grid represents a certain region in the real world. Based on the LiDAR point clouds $P$ and pre-defined BEV grids, we first divide $P$ into several bins $\{b_1, b_2, ..., b_{n^2}\}$, where $n$ is the number of grids on the BEV edge. Each bin contains an indefinite number of points $\{p^1, p^2, ..., p^l\}$ and corresponds to a certain BEV grid. For each bin, we project these points to the image space, retaining only the points that fall onto the image plane, and use the corresponding pixel features in the HR feature as the features of the valid points. Then, we aggregate the features of these valid points and let the aggregated feature be the feature of the corresponding BEV grid. We repeat the above operation for each bin and obtain the point BEV. However, the LiDAR points are sparse and not every point can be projected to the image plane, so the valid grids in point BEV are sparse. To tackle this problem, we fuse the point BEV with dense ray BEV to get the final camera BEV. In addition, from the pixel coordinate to the 3D coordinate, point-based transformation can be equivalently converted into calculating the depth of the pixels owning correspondences with points and then projecting these pixels and their corresponding features into 3D space. Therefore, this process also uses depth information. Compared with the ray-based transformation, point-based transformation only
handles pixels owning the correspondences with points, while the ray-based transformation deals with each pixel based on depth prediction. 
The point-based transformation provides an accurate and fast view transformation, while the ray-based transformation helps to supplement the space that is not scanned by LiDAR with 2D features.

\begin{figure*}[t]%% placement specifier
\centering
\includegraphics[scale=0.45]{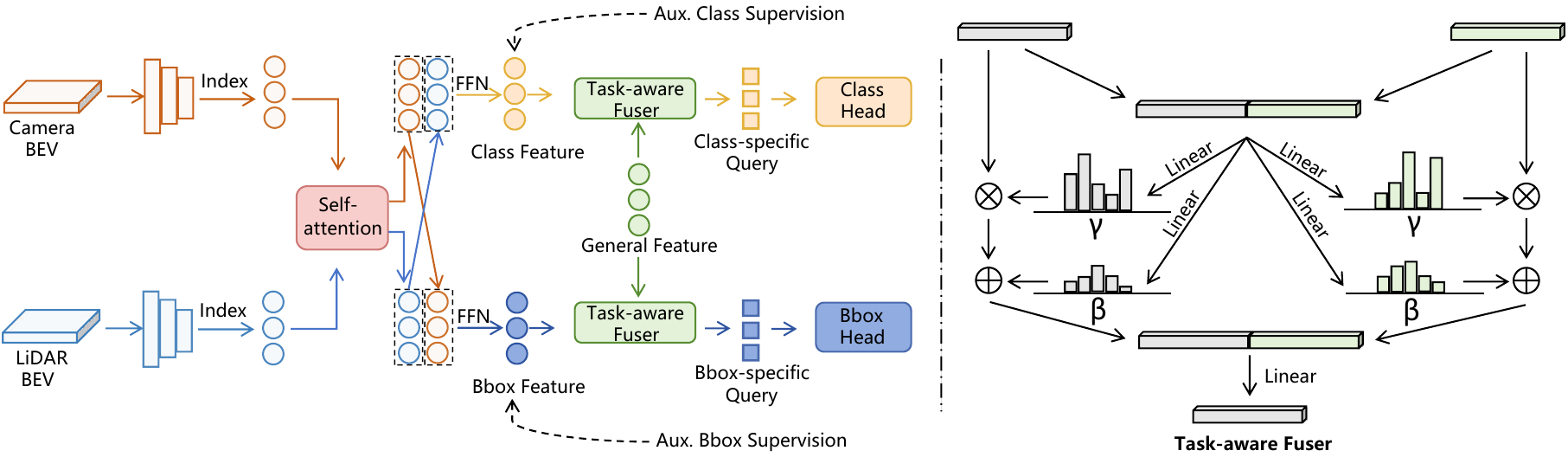}
\caption{The overall framework of our proposed task-specific predictor.}\label{fig4}
\end{figure*}

\subsection{Task-Specific Predictor}\label{sec:3.3}
The overall framework of our proposed task-specific predictor is shown in Fig. \ref{fig4}. The purpose of the task-specific predictor is to tackle the mutual suppression between classification prediction and bounding box regression during joint training. To this end, we explicitly and controllably construct task-specific features and integrate them with general features to build task-specific queries. Sub-task heads will use the corresponding task-specific queries to predict the certain attributes of the objects.

\subsubsection{Task-Specific Feature}\label{sec:3.3.1}

Given camera BEV $B_c$ and LiDAR BEV $B_l$, we first use CNN-based encoders to obtain specific features, and the parameters of these two specific encoders are not shared. The outputs of these encoders are denoted as $ \widetilde{B_c}$ and $ \widetilde{B_l}$. To maintain the consistency of the candidate position and facilitate the subsequent fusion between general and task-specific features, we directly use the candidate indexes $I_p$ generated in Sec. \ref{sec:3.1} to select candidates in specific feature extractor. Given the $\widetilde{B_c}$, $\widetilde{B_l}$, and the indexes of candidates $I_p$, we first generate camera candidates $Q_c=\{Q_{c,k}\}_{k=1}^K$ from $\widetilde{B_c}$ and $I_p$. We then generate LiDAR candidates $Q_l=\{Q_{l,k}\}_{k=1}^K$ from $\widetilde{B_l}$ and $I_p$. After that, we concatenate $Q_c$ and $Q_l$ together and use a standard self-attention module to fuse these two modal candidates. Then, the corresponding LiDAR and camera candidates are fused by the feed-forward network (FFN) to obtain the class feature $f_c$ and Bbox feature $f_b$. 

\subsubsection{Task-Specific Supervision}

Although the method in Sec. \ref{sec:3.3.1} explicitly constructs the task-specific features, it cannot guarantee the correlation between these features and their corresponding sub-tasks. In other words, we hope that the sub-task preferences of these features are controllable. The best way of achieving this goal is to use supervision which makes class features focus on classification prediction and Bbox features focus on bounding box prediction. In view of this, we propose task-specific supervision which is only conducted in the training stage.

We use a feed-forward network (FFN) to construct a class auxiliary task head and a Bbox auxiliary task head. Given class feature $f_c$ and Bbox feature $f_b$, $f_c$ is used to predict class, and $f_b$ is used to predict Bbox. The class predicted result and its corresponding Bbox predicted result will be combined as an auxiliary predicted result $o$ for one object. We use bipartite matching \cite{carion2020end} between the predictions and ground-truth objects $\hat{o}$ through the Hungarian algorithm. We calculate the auxiliary loss based on Eq. \ref{eq2}. 
\begin{equation}\label{eq2}
L_{aux}(o, \hat{o}) = \lambda_1L_{cls}(c, \hat{c}) + \lambda_2L_{Bbox}(b, \hat{b}),
\end{equation}
where $\lambda_1$ and $\lambda_2$ are the hyper-parameters to balance class-aux loss function $L_{cls}$ and Bbox-aux loss function $L_{Bbox}$, c, b are the predicted object’s category and 3D bounding box, and $\hat{c}$, $\hat{b}$ denote the corresponding ground-truth. We use focal loss \cite{chen2022focal} as $L_{cls}$ and L1 Loss as $L_{Bbox}$.

\subsubsection{Task-Specific Fuser}
Given the general feature and the task-specific feature, we use a task-specific fuser to adaptively fuse the specific and general features, which can select different information of interest for different sub-tasks. We first concatenate general and specific features. Then, we predict the attention vector $\gamma$ and $\beta$ via Eq. 3.
\begin{equation}\label{eq3}
\gamma = \psi_1([f_g, f_s]), \beta = \psi_1([f_g, f_s]),
\end{equation}
where $\psi_{i=1,2}(.)$ are the linear mapping functions, $[.,.]$ is the concatenate function and $f_s$ is the task-specific feature.
We predict a total of two groups $\gamma$ and $\beta$, one of which is used to adjust the information preference for specific features, and the other is used to adjust the information preference for general features. After that, we obtain the task-specific query $Q_s$ via Eq. \ref{eq4}.
\begin{equation}\label{eq4}
Q_s = \psi([\gamma_s \times f_s + \beta_s, \gamma_g \times f_g + \beta_g]),
\end{equation}
where $\psi(.)$ is the linear function, $[.,.]$ is the concatenate function, $[\gamma_s, \beta_s]$ and  $[\gamma_g, \beta_g]$ are the attention vectors corresponding to task-specific and general feature, respectively. Finally, the task-specific query will be inputted to the sub-task head for predicted 3D objects.

\section{Experiments}
\subsection{Dataset and metrics}

We conduct extensive experiments on a large-scale autonomous driving dataset nuScenes \cite{caesar2020nuscenes}, which is widely used for 3D LiDAR-camera object detection and 3D multi-object tracking task. 
% This dataset has a total of 1000 scenes and is officially divided into 700/150/150 for training/validation/testing. For each frame, nuScenes has 6 images and 1 point cloud from 32 beams LiDAR. The multi-view camera group has the same field of view as LiDAR point cloud, $360^{\circ}$. There are around 1.4 million annotated 3D bounding boxes for ten classes.
There are a total of 10 classes of objects and around 1.4 million annotated 3D bounding boxes in nuScenes. NuScenes totally has 1000 scenes in different conditions, such as sunny, rainy, day, and night. For each frame, nuScenes has 6 images and 1 point cloud from 32 beams LiDAR. The multi-view camera group has the same field-of-view as LiDAR point clouds, $360^{\circ}$. We follow the official instruction to divide 1000 scenes into 700/150/150 for training/validation/testing. Following common practice, we transform the points from the previous 9 frames to the current frame for training and evaluation.
For 3D object detection, 
we follow BEVFusion \cite{liu2023bevfusion} to use nuScenes detection score (NDS) and mean average precision (mAP) as the metrics. Unlike 2D detection, the mAP is defined by the distance of the center instead of the Intersection over Union (IOU). The final mAP is the average result of distances 0.5m, 1m, 2m, and 4m among ten classes. For NDS, it is the weighted sum of mAP and other official predefined metrics
including average translation error (ATE), average scale error (ASE), average orientation error (AOE), average velocity error (AVE), average attribute error (AAE). For the 3D multi-object tracking task, we follow Centerpoint \cite{yin2021center} to use average multi-object tracking accuracy (AMOTA), average multi-object tracking precision (AMOTP), and ID switch as the metrics.

\subsection{Implement Details}

\begin{table*}[h!]
    \caption{Detection performance comparison on nuScenes test set.‘C.V.’, ‘T.L.’, ‘B.R.’, ‘M.T.’, ‘Ped.’, and ‘T.C.’ indicate the construction vehicle, trailer, barrier, motorcycle, pedestrian, and a traffic cone, respectively. The best performance is highlighted in bold, while the second best is marked underlined. Results are the performance of the single model without any test-time augmentation or model ensemble.}
    \centering
    \label{tab:table1}
    \small
    \setlength{\tabcolsep}{0.2mm}
    \begin{tabular}{c|cccccccccc|cc}
    \hline
    Method&Car&Truck&C.V.&Bus&T.L.&B.R.&M.T.&Bike&Ped.&T.C.&mAP&NDS\\
    \hline
    \multicolumn{13}{c}{LiDAR Methods} \\
    \hline
    CenterPoint\cite{yin2021center}&84.6&51.0&17.5&60.2&53.2&70.9&53.7&28.7&83.4&76.7&58.0&65.5\\
    VoxelNetXt\cite{chen2023voxelnext}&84.6&53.0&28.7&64.7&55.8&74.6&73.2&45.7&85.8&79.0&64.5&70.0\\
    TransFusion-L\cite{bai2022transfusion}&86.2&56.7&28.2&66.3&58.8&78.2&68.3&44.2&86.1&82.0&65.5&70.2\\
    HEDNet\cite{zhang2024hednet}&87.1&56.5&33.6&70.4&63.5&78.1&70.4&44.8&87.9&85.1&67.7&72.0\\
    FocalFormer3D\cite{chen2023focalformer3d}&87.2&57.1&34.4&69.6&64.9&77.8&76.2&49.6&88.2&82.3&68.7&72.6
    \\
    DSVT \cite{wang2023dsvt}&-&-&-&-&-&-&-&-&-&-&68.4&72.7\\
    LION\cite{liu2024lion}&87.1&61.1&36.3&68.9&65.0&79.5&74.0&49.2&90.0&87.3&69.8&73.9\\
    \hline
    \multicolumn{13}{c}{LiDAR-Camera Methods} \\
    \hline
    GraphAlign\cite{song2023graphalign}&87.6&57.7&26.1&66.2&57.8&74.1&72.5&49.0&87.2&86.3&66.5&70.6\\
    AutoAlign\cite{chen2022autoalign}&85.9&55.3&29.6&67.7&55.6&-&71.5&51.5&86.4&-&65.8&70.9\\
    
    UVTR\cite{li2022unifying} &87.5&56.0&33.8&67.5&59.5&73.0&73.4&54.8&86.3&79.6&67.1&71.1\\
    TansFusion\cite{bai2022transfusion}&87.1&60.0&33.1&68.3&60.8&78.1&73.6&52.9&88.4&86.7&68.9&71.7\\
    BEVFusion(PKU)\cite{liang2022bevfusion}&88.1&60.9&34.4&69.3&62.1&78.2&72.2&52.2&89.2&85.5&69.2&71.8\\

    GraphAlign++\cite{song2023graphalign++}&87.5&58.5&32.3&68.9&58.3&74.3&76.4&53.9&88.3&86.3&68.5&72.2\\
    
    Autoalignv2\cite{chen2022deformable}&87.0&59.0&33.1&69.3&59.3&-&72.9&52.1&87.6&-&68.4&72.4\\
    BEVFusion(MIT)\cite{liu2023bevfusion} &88.6&60.1&39.3&69.8&63.8&80.0&74.1&51.0&89.2&86.5&70.2&72.9\\
    ObjectFusion\cite{cai2023objectfusion}&89.4&59.0&40.5&71.8&63.1&80.0&78.1&53.2&90.7&87.7&71.0&73.3\\
    DeepInteraction\cite{yang2022deepinteraction}&87.9&60.2&37.5&70.8&63.8&80.4&75.4&54.5&91.7&87.2&70.8&73.4\\
    GraphBEV\cite{song2024graphbev}&89.2&60.0&40.8&72.1&64.5&80.1&76.8&53.3&90.9&88.9&71.7&73.6\\
    ContrastAlign\cite{song2024contrastalign}&89.0&60.9&41.1&70.1&64.6&82.2&75.9&53.8&90.9&89.3&71.8&73.8\\
    SparseFusion\cite{xie2023sparsefusion}&88.0&60.2&38.7&72.0&64.9&79.2&78.5&59.8&90.9&87.9&72.0&73.8\\
    FocalFormer3D-F\cite{chen2023focalformer3d}&88.5&61.4&35.9&71.7&66.4&79.3&80.3&57.1&89.7&85.3&71.6&73.9\\
    MSMDFusion\cite{jiao2023msmdfusion}&88.4&61.0&35.2&71.4&64.2&80.7&76.9&58.3&90.6&88.1&71.5&74.0\\
    CMT\cite{yan2023cross}&88.0&63.3&37.3&75.4&65.4&78.2&79.1&60.6&87.9&84.7&72.0&74.1\\
    UniTR\cite{wang2023unitr}&87.9&60.2&39.3&72.2&65.1&76.8&75.8&52.3&89.4&89.7&70.9&74.5\\
    DAL\cite{huang2023detecting}&-&-&-&-&-&-&-&-&-&-&72.0&74.8\\
    
    IS-Fusion\cite{yin2024fusion}&88.3&62.7&38.4&74.9&67.3&78.1&82.4&59.5&89.3&89.2&\underline{73.0}&\underline{75.2}\\
    \rowcolor{Gray}CoreNet(ours)&89.7&63.4&39.9&73.9&66.7&79.1&80.3&59.3&91.7&88.5&\textbf{73.3}&\textbf{75.6}\\
    \hline
    \end{tabular}
\end{table*}

We implement our network based on MMDetection3D \cite{mmdet3d2020} and BEVFusion \cite{liu2023bevfusion} codebase. For multi-object tracking tasks, we adopt the tracking test script in the Centerpoint\cite{yin2021center} codebase to provide a fair comparison. We adopt Swin-T \cite{liu2021swin} as our image encoder and VoxelNet \cite{zhou2018voxelnet} as our LiDAR encoder that is mentioned in Sec. \ref{sec:3.1}. We first pre-train the image encoder and LiDAR encoder, following \cite{liu2023bevfusion,huang2023detecting}. Then, we train CoreNet for another 6 epochs based on these two pre-trained encoders. Following previous methods \cite{liu2023bevfusion,liang2022bevfusion}, we also use CBGS \cite{zhu2019class} in the pre-train and fusion processes. The voxel size is set to (0.075m, 0.075m, 0.2m), and the BEV feature map is set to 180$\times$180. During the CoreNet training, we use the same training strategy and data augmentation as BEVFusion \cite{liu2023bevfusion}. Specifically, the learning rate is adjusted by cycle policy with the initial value 2.0$\times$$10^{-4}$. We conduct optimization based on AdamW with weight decay value $10^{-2}$. During the training process, we don't freeze any pre-trained backbone, and the entire model is trained in an end-to-end manner. All the codes are implemented by PyTorch, and the model is trained with a batch size of 16 on 8 NVIDIA A100 GPUs. For the evaluation results, we only show the performance of the single model without any test-time augmentation or model ensemble.

\subsection{Performance Comparison}

\subsubsection{Detection results comparison} 

We compare CoreNet with other existing approaches on nuScenes \cite{caesar2020nuscenes} test set as shown in Table \ref{tab:table1}. CoreNet performs better than all the other single-modality and multi-modality methods by achieving 73.3\% mAP and 75.6\% NDS. Compared to BEVFusion(MIT) \cite{liu2023bevfusion} which is the baseline, CoreNet improves the results by 2.7\% NDS and 3.1\% mAP. Compared to recent SOTA methods, IS-Fusion \cite{yin2024fusion} and DAL \cite{huang2023detecting}, CoreNet also achieves better results. It should be noted that CoreNet is a flexible model. The dual-stream transformation and task-specific predictor proposed by CoreNet can be extracted and combined with other methods. 
% We also compare the detection performance on nuScenes validation set. As shown in Table \ref{tab:table2},
% CoreNet gets the best performance (74.5\% NDS, 72.6\% mAP). In particular, it surpasses the SOTA methods on different sizes of image inputs. 

\begin{figure*}[t]%% placement specifier
\centering
\includegraphics[scale=0.47]{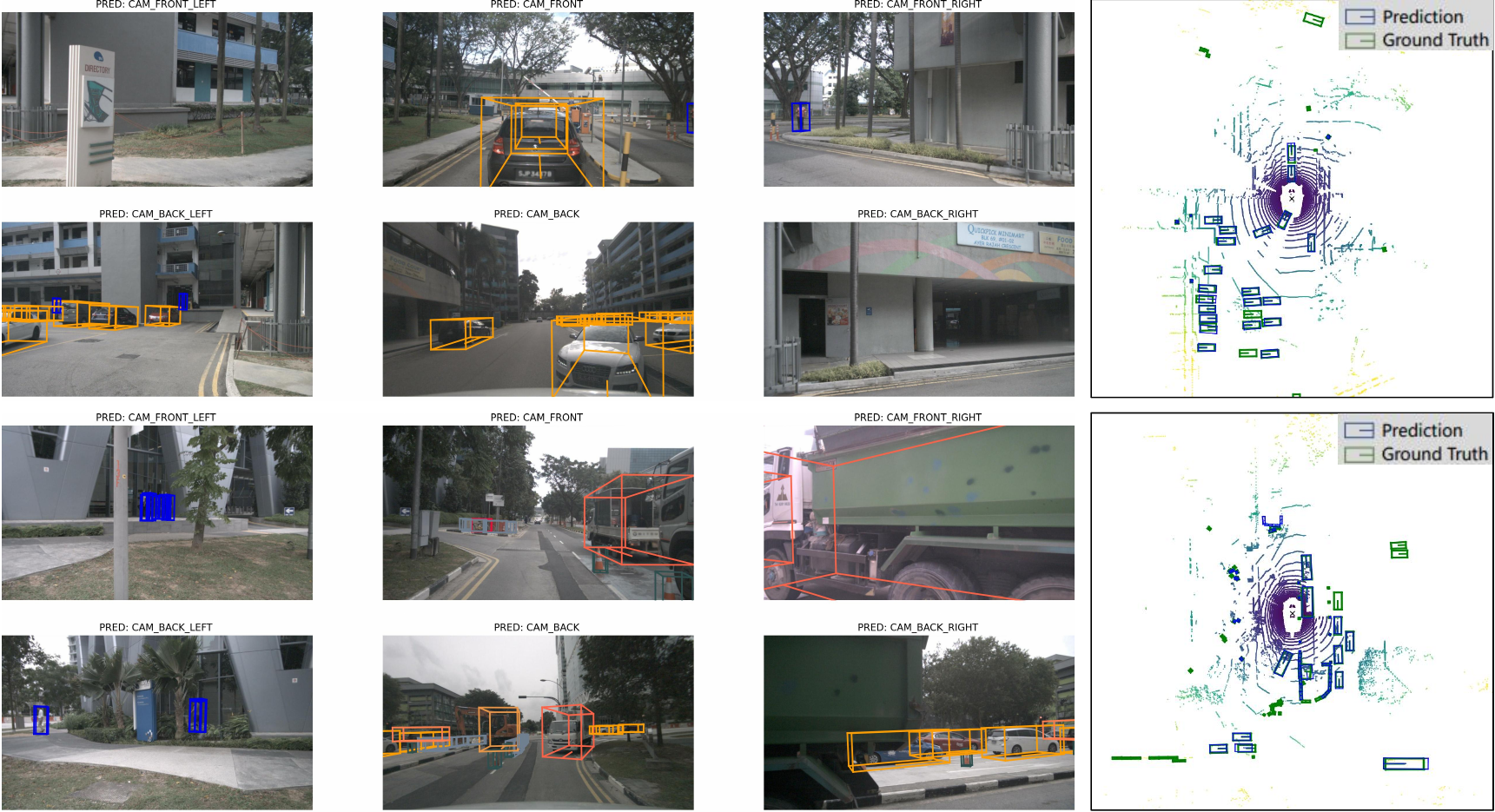}
\caption{The visualization of our proposed method on nuScenes validation set.}\label{fig5}

\end{figure*}
\begin{table*}[h!]
  \caption{Detection performance comparison on nuScenes validation set. The best performance is highlighted in bold, while the second best is marked underlined.}
  \small
  \setlength{\tabcolsep}{0.5mm}
  \label{tab:table2}
  \centering
  \begin{tabular}{c|ccc|cc}
    \hline
     Method & Image encoder & Voxel encoder &Image size& mAP(\%)$\uparrow$ & NDS(\%)$\uparrow$\\
     \hline
      CMT \cite{yan2023cross} & ResNet-50 \cite{he2016deep} & VoxelNet \cite{zhou2018voxelnet} &320$\times$800& 67.9 & 70.8\\
      Transfusion \cite{bai2022transfusion} & ResNet-50 \cite{he2016deep} & VoxelNet \cite{zhou2018voxelnet} & 448$\times$800& 68.5 & 71.4\\
     BEVFusion \cite{liu2023bevfusion} & Swin-T \cite{he2016deep} & VoxelNet \cite{zhou2018voxelnet} & 256$\times$704& 68.5 & 71.4\\
     DeepInteraction \cite{yang2022deepinteraction} & ResNet-50 \cite{he2016deep} & VoxelNet \cite{zhou2018voxelnet} & 448$\times$800& 69.9 & 72.6\\
    GraphBEV \cite{song2024graphbev} & Swin-T \cite{he2016deep} & VoxelNet \cite{zhou2018voxelnet} & 256$\times$704& 70.1 & 72.9\\
     SparseFusion \cite{xie2023sparsefusion} & Swin-T \cite{liu2021swin} & VoxelNet \cite{zhou2018voxelnet} &448$\times$800& 71.0 & 73.1\\
     UniTR \cite{wang2023unitr} & DSVT \cite{wang2023dsvt} & DSVT \cite{wang2023dsvt} &256$\times$704& 70.5 & 73.3\\
     DAL \cite{huang2023detecting}& ResNet-50 \cite{he2016deep} & VoxelNet \cite{zhou2018voxelnet}&256$\times$704 & 70.0 & 73.4\\
     IS-Fusion \cite{yin2024fusion} & Swin-T \cite{liu2021swin} & VoxelNet \cite{zhou2018voxelnet} &256$\times$704& 72.4 & 73.7\\
     \rowcolor{Gray}CoreNet (ours)& Swin-T \cite{liu2021swin} & VoxelNet \cite{zhou2018voxelnet}&256$\times$704 & 71.5 & 73.9\\
     \hline
      CMT \cite{yan2023cross} & VoVNet \cite{lee2020centermask} & VoxelNet \cite{zhou2018voxelnet} &640$\times$1600& 70.3 & 72.9\\
      \rowcolor{Gray}CoreNet (ours)& VoVNet \cite{lee2020centermask} & VoxelNet \cite{zhou2018voxelnet}&640$\times$1600 &72.3 & 74.2 \\
     DAL \cite{huang2023detecting}& ResNet-50 \cite{he2016deep} & VoxelNet \cite{zhou2018voxelnet}&384$\times$1056 & 71.5 & 74.0\\
     \rowcolor{Gray}CoreNet (ours)& ResNet-50 \cite{he2016deep} & VoxelNet \cite{zhou2018voxelnet}&384$\times$1056 & 71.7& 73.9\\
     IS-Fusion \cite{yin2024fusion} & Swin-T \cite{liu2021swin} & VoxelNet \cite{zhou2018voxelnet} &384$\times$1056& \textbf{72.8} & \underline{74.0}\\
     \rowcolor{Gray} CoreNet (ours)& Swin-T \cite{liu2021swin} & VoxelNet \cite{zhou2018voxelnet}&384$\times$1056 & \underline{72.6} & \textbf{74.5}\\
  \hline
  \end{tabular}
\end{table*}

\begin{figure*}[h!]%% placement specifier
\centering
\includegraphics[scale=0.8]{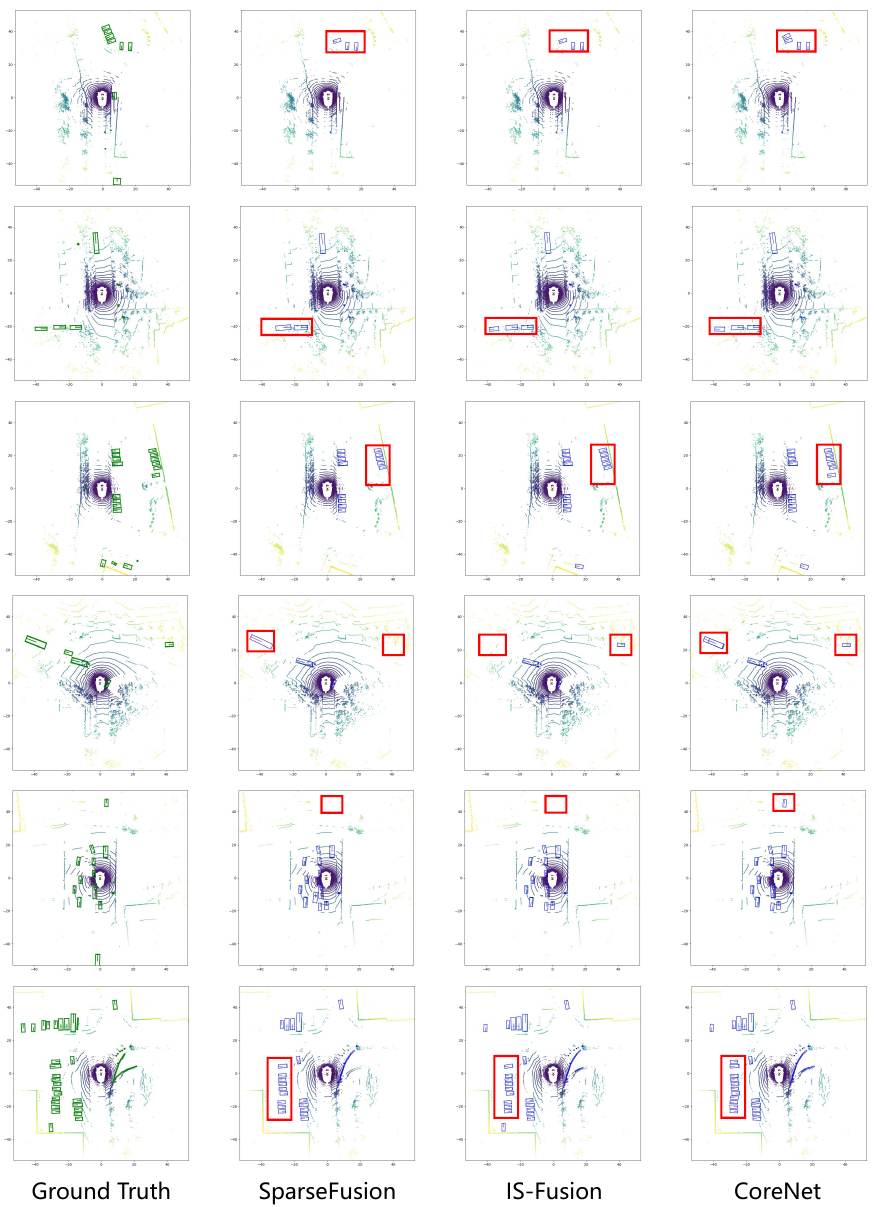}
\caption{The visualization of our proposed method on nuScenes validation set. The red box indicates the areas where we successfully predicted but previous methods failed.}\label{fig6}
\end{figure*}

As shown in Table \ref{tab:table2}, we compare the detection performance on the nuScenes validation set and achieve the best performance by achieving 74.5\% NDS and 72.6\% mAP. We find that CoreNet surpasses the SOTA methods on different sizes of image inputs. Compared with different backbones including ResNet-50 \cite{he2016deep} and VoVNet \cite{lee2020centermask}, CoreNet can also achieve satisfactory results. As shown in Table \ref{tab:speed}, we compare the inference speed and parameter size between CoreNet and the existing SOTA IS-Fusion. It could be observed that CoreNet has a faster inference speed.
We visualize some qualitative results of CoreNet to prove its superiority, as shown in Fig. \ref{fig5}. We also compare the visualization among ground-truth, SparseFusion, IS-Fusion, and CoreNet in Fig. \ref{fig6}. Our proposed model shows strong competitiveness in 3D object detection tasks, which fully demonstrates the following viewpoint: reducing the depth ambiguity in 2D-to-BEV transformation and alleviating suppression between different sub-tasks can boost performance. 

% \clearpage

\begin{table*}[h!]
  \caption{Comparison of inference speed and parameter size on nuScenes validation set with Swin-T backbones. We test all the models and report results on a single NVIDIA Tesla A100 GPU without considering the time cost of voxelization in the voxel backbone. The size of input images is set to 384$\times$1056.}
  \small
  \setlength{\tabcolsep}{0.5mm}
  \label{tab:speed}
  \centering
  \begin{tabular}
  % {cc|ccc|cccc}
  {
    p{70pt}<{\centering}p{45pt}<{\centering}p{60pt}<{\centering}
    p{40pt}<{\centering}p{40pt}<{\centering}p{35pt}<{\centering} 
    p{50pt}<{\centering}p{40pt}<{\centering}
    % p{20pt}<{\centering}  
  } 
  \hline
  \multirow{2}{*}{Method}&\multicolumn{4}{c}{Latency(ms)}&\multirow{2}{*}{FPS$\uparrow$}&\multirow{2}{*}{\#parm.}&\multirow{2}{*}{NDS(\%)}\\
  &Backbone&ViewTrans&Encoder&Decoder&&&\\
  \hline
  IS-Fusion\cite{yin2024fusion}&87&35&193&8&3.1&53.4M&74.0\\
  CoreNet&74&110&22&15&4.5&87.5M&74.5\\
  \hline
  \end{tabular}
\end{table*}

\subsubsection{3D multi-object tracking comparison}
\begin{table*}[h]
  \caption{Tracking performance comparison on nuScenes validation set. The Modality column: “L” means only use LiDAR data; “LC” means use both LiDAR and camera data. The best performance is highlighted in bold, while the second best is marked underlined.}
  \label{tab:table3}
  \small
  \setlength{\tabcolsep}{0.5mm}
  \centering
  \begin{tabular}
  % {cc|ccc}
  {p{100pt}<{\centering}p{50pt}<{\centering}p{75pt}<{\centering}
    p{75pt}<{\centering}p{70pt}<{\centering}}
    \hline
    Method & Modality & AMOTA (\%) $\uparrow$ & AMOTP (\%) $\downarrow$ & ID switch $\downarrow$\\
    \hline 
    Centerpoint \cite{yin2021center} & L & 63.7 & 60.6 & 640 \\
    TransFusion-L \cite{bai2022transfusion} & L & 69.9 & 59.9 & 821 \\
    TransFusion \cite{bai2022transfusion} & LC & 71.8 & 60.3 & 794 \\
    BEVFusion \cite{liu2023bevfusion} & LC & 72.8 & 59.4 & 764 \\
    ObjectFusion \cite{cai2023objectfusion} & LC & \underline{74.2} & 54.3 & \textbf{611} \\
    SparseFusion \cite{xie2023sparsefusion} & LC & 61.1 & 56.0 & 930 \\
    IS-Fusion \cite{yin2024fusion} & LC & 73.6 & \underline{52.7} & 1017 \\
    \textbf{CoreNet (ours) }& LC & \textbf{76.3}  & \textbf{46.3} & \underline{656}\\
  \hline
  \end{tabular}
\end{table*}

We also conduct 3D multi-object tracking (MOT) experiments on the nuScenes tracking benchmark. Following previous works \cite{bai2022transfusion, cai2023objectfusion}, we use the same tracking-by-detection algorithm, which directly connects objects in consecutive frames based on the greedy algorithm.
As shown in Table \ref{tab:table3}, our proposed CoreNet achieves competitive results (76.3\% in AMOTA, 46.3\% in AMOTP, 656 in ID switch). Compared to our baseline BEVFusion \cite{liu2023bevfusion}, CoreNet improves AMOTA by 3.5\%. Compared to IS-Fusion \cite{yin2024fusion} which shows good performance in 3D detection task, CoreNet improves AMOTA by 2.7\%. 
% This phenomenon proves the considerable ability of CoreNet in 3D MOT task. 
This indicates that the detection results of CoreNet have better stability and continuity, and CoreNet could be a strong solution for the 3D MOT task.

\subsection{Ablation Study}
In this section, we investigate the effectiveness of each proposed component. Our ablation study is conducted on nuScenes validation set by using Swin-T \cite{liu2021swin} backbone and setting image size to 256$\times$604.  For the baseline of our method, we have two changes compared to the original BEVFusion \cite{liu2023bevfusion}: (1) we use velocity augment \cite{huang2023detecting} to increase the precision of velocity prediction and (2) we change the original shallow BEV Fuser into a sophisticated CNN-based network. As shown in Table \ref{tab4}, we first evaluate the performance of the two main modules, i.e., dual-stream transformation (DST) and task-specific predictor (TSP). DST improves the results by 0.7\% mAP and 0.9\% NDS, while TSP improves the results by 1.1\% mAP and 1.3\% NDS. Adopting DST and TSP simultaneously can improve the results by 2.8\% mAP and 1.9\% NDS, demonstrating the effect of tackling depth ambiguity and sub-task suppression.

\begin{table*}[t!] 
  \caption{Ablation study for each module in CoreNet. DST indicates dual-stream transformation, and TSP indicates task-specific predictor.}
  \small
  \label{tab4}
  \centering
  \begin{tabular}{cc|ccccc|cc}
    \hline
    DST & TSP & mATE & mASE & mAOE & mAVE & mAAE & mAP & NDS\\
    \hline 
    &&0.278 &	0.256 &	0.319 &	0.201 &	0.182 &  68.7 & 72.0\\
     \checkmark&  & 0.274  & 0.254 & 0.284 & 0.187 & 0.184 & 69.4
 & 72.9\\
     &  \checkmark& 0.274 & 0.253 & 0.268 & 0.184 & 0.182 & 69.8
 & 73.3\\
     \checkmark&  \checkmark& 0.273 & 0.254 & 0.287 & 0.187 & 0.183 & \textbf{71.5}
 & \textbf{73.9}
\\
  \hline
  \end{tabular}
\end{table*}

\begin{table*}[t!] 
  \caption{Component-wise ablation study on dual-stream transformation.}
  \small
  \label{tab5}
  \centering
  \begin{tabular}{cc|ccccc|cc}
    \hline
    Ray & Point & mATE & mASE & mAOE & mAVE & mAAE & mAP & NDS\\
    \hline 
     \checkmark&  & 0.275 & 0.254 & 0.307 & 0.197 & 0.184 & 68.9
 & 72.3\\
     &  \checkmark& 0.270 & 0.254 & 0.300 & 0.200 & 0.185& 
 69.1 & 72.5\\
     \checkmark&  \checkmark& 0.274  & 0.254 & 0.284 & 0.187 & 0.184 & \textbf{69.4}
 & \textbf{72.9}
\\
  \hline
  \end{tabular}
\end{table*}

\begin{table*}[t!] 
  \caption{Component-wise ablation study on task-specific predictor. TS is the abbreviation of task-specific.}
  \small
  \label{tab6}
  \centering
  \begin{tabular}{c|ccccc|cc}
    \hline
    component & mATE & mASE & mAOE & mAVE & mAAE & mAP & NDS\\
    \hline 
    + TS feature
 & 0.278  & 0.256 & 0.297 & 0.192 & 0.183 & 
 69.2& 72.5\\
   + TS supervision
 & 0.277  & 0.253 & 0.291 & 0.189 & 0.181 &
 69.6 & 72.9 \\
   + TS fuser & 0.274 & 0.253 & 0.268 & 0.184 & 0.182 & \textbf{69.8}
 & \textbf{73.3}
\\
  \hline
  \end{tabular}
\end{table*}

\begin{figure*}[t!]%% placement specifier
\centering
\includegraphics[scale=0.5]{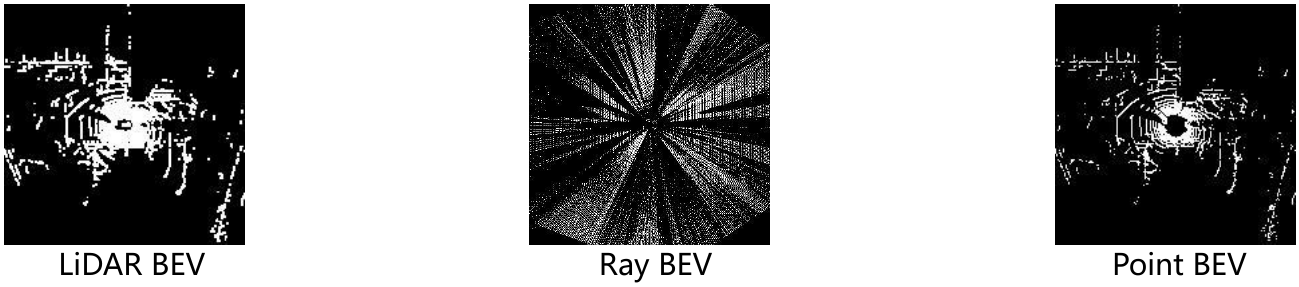}
\caption{Visualization of the ray BEV and the point BEV. Bright grids indicate the presence of features here.}\label{fig7}
\end{figure*}

The dual-stream transformation consists of two main streams, i.e., ray-based and point-based streams. The purpose of these two streams is both to alleviate the depth ambiguity caused by point-pixel misalignment. As shown in Table \ref{tab5}, we investigate the effectiveness of these two streams. Ray-based transformation improves the results by mAP 0.2\% and 0.3\% NDS, while point-based stream improves the results by 0.4\% mAP and 0.5\% NDS. This result highlights the crucial role of projecting 2D pixel features into 3D points should comply with occlusion relationships. 
As shown in Fig. \ref{fig7}, we visualize the ray BEV and the point BEV.
It can be observed that the number of valid grids of ray BEV is higher, while the valid grid distribution of point BEV is closer to that of LiDAR. It should be noted that ray BEV also generates multiple features on each pixel-object ray, but through one-hot supervision, pixel features are mainly concentrated on a certain point.
The motivation of task-specific predictor is to relieve the mutual suppression between class and Bbox prediction by constructing a task-specific query. It consists of three key parts: (1) task-specific feature, (2) task-specific supervision, and (3) task-specific fuser. As shown in Table \ref{tab6}, we use a step-wise way to evaluate their performance. All these three components contribute to the final results. The full task-specific predictor improves the results by 1.1\% mAP and 1.3\% NDS. This phenomenon highlights the importance of reducing mutual suppression across different sub-tasks. Additionally, these two components can easily be inserted into other BEV-based methods for further application.
\subsection{Robustness Analysis}

\begin{table*}[t!] 
  \caption{Robustness analysis on nuScenes validation
set with different lighting and weather conditions. We use mAP(\%) as the metric for evaluation.}
  \small
  \label{tab7}
  \centering
  \begin{tabular}{cc|cccc}
    \hline
    Method & Modality & Sunny & Rainy & Day & Night\\
    \hline 
    CenterPoint \cite{yin2021center} &L&62.9&59.2&62.8&35.4\\
    BEVDet \cite{huang2021bevdet} &C&32.9&33.7&33.7&13.5\\
    BEVFusion \cite{liu2023bevfusion} &LC&68.2&69.9&68.5&42.8 \\
    GraphBEV \cite{song2024graphbev} &LC&70.1&70.2&69.7&45.1 \\
    SparseFusion \cite{xie2023sparsefusion} &LC&70.3&70.4&70.5&\underline{44.7} \\
    IS-Fusion \cite{yin2024fusion} &LC&\textbf{72.2}&\underline{71.7}&\underline{72.5}&44.4\\
    CoreNet (Ours) &LC&\textbf{72.2}&\textbf{73.6}&\textbf{72.6}&\textbf{45.0}\\
  \hline
  \end{tabular}
\end{table*}

In this section, we conduct a robustness analysis to simulate the outside environment in real-world usage scenarios. The robustness is measured under different conditions, including lighting, weather conditions, object distances, and sizes. All these experiments are conducted on the nuScenes validation set.

We first evaluate the performance under lighting and weather changes, which are challenging as they can affect the quality of data received by sensors. Following BEVFusion \cite{liu2023bevfusion}, we split the scenes into Sunny, Rainy, Day, and Night by searching the keywords 'rain' and 'night' in the scene description. As shown in Table \ref{tab7}, our proposed CoreNet surpasses the previous methods under different conditions. Compared to the baseline BEVFusion, CoreNet improves mAP by 4.0\%, 3.7\%, 4.1\%, and 2.2\% on the Sunny, Rainy, Day, and Night conditions, respectively. Compared to the SOTA method IS-Fusion, CoreNet improves mAP by 1.9 \% in the rainy condition. This indicates that our method has better stability in scenarios with noise.

\begin{table*}[t!] 
  \caption{Robustness analysis on nuScenes validation
set with different ego distances and object sizes. We use mAP(\%) as the metric for evaluation.}
  \small
  \label{tab8}
  \centering
  \begin{tabular}{cc|ccc|cc}
    \hline
    Method & Modality & Near & Middle & Far & Small & Large\\
    \hline 
    TransFusion-L \cite{bai2022transfusion} & L & 77.1 & 60.4 & 43.2 & 56.1 & 53.1\\
    BEVFusion \cite{liu2023bevfusion} & LC & 78.5 & 64.6 & 50.5 & 61.9 & 57.9\\
    GraphBEV \cite{song2024graphbev} & LC & 78.6& 65.3 & 42.1 & - & -\\
    SparseFusion \cite{xie2023sparsefusion} & LC & 81.3 & 66.8 & 51.6 & 64.8 & 58.2\\
    
    IS-Fusion \cite{yin2024fusion} & LC & \underline{81.5} & \textbf{69.8} & \underline{55.6} & \textbf{65.9} & \underline{62.0}\\
    CoreNet (Ours) & LC & \textbf{82.2} & \underline{68.3} & \textbf{56.3} & \underline{65.7} & \textbf{62.2}\\
  \hline
  \end{tabular}
\end{table*}

We then analyze the performance under different distances and sizes. Following BEVFusion \cite{liu2023bevfusion}, we classify objects into three categories based on their distance to the vehicle: Near (0-20m), Middle (20-30m), and Far ($>$30m). We also divide the objects into Small (0-4m) and Large ($>$4m) groups according to their sizes. As shown in Table \ref{tab8}, compared to the baseline and latest SOTA methods \cite{yin2024fusion, xie2023sparsefusion} CoreNet achieves the best performance. In the hard case, i.e., detecting objects at a long distance, CoreNet surpasses BEVFusion and IS-Fusion by 5.8\% and 0.7\% mAP, respectively. Long distances will reduce the useful information captured by sensors. Better results in this scene indicate that when there is insufficient effective information about objects, the prediction results of CoreNet are more robust.

\begin{table*}[!t]
\footnotesize
\setlength{\tabcolsep}{1mm}
\caption{Comparison of robust results under different corruptions on nuScenes-C \cite{dong2023benchmarking} benchmark. 
The original performance, the performance under each corruption and the overall averaged corruption robustness $\mathrm{mAP_{cor}}$ are shown.}
\label{tab:10}
\begin{center}
\resizebox{\textwidth}{!}{ 
\begin{tabular}{c|c|ccc|ccc|cccc}
\hline
\multicolumn{2}{c|}{\multirow{2}{*}{\textbf{Corruption}}} & \multicolumn{3}{c|}{\textbf{LiDAR-only}} &  \multicolumn{3}{c|}{\textbf{Camera-only}} & \multicolumn{3}{c}{\textbf{LC Fusion}} \\
\multicolumn{2}{c|}{} & PointPillars & SSN & CenterPoint & PGD & DETR3D & Bevformer & FUTR3D & TransFusion & BEVFusion & Ours\\
\hline
\rowcolor{Gray}\multicolumn{2}{c|}{\textbf{None} ($\mathrm{mAP_{clean}}$)} & 27.69 & 46.65 & \bf59.28 & 23.19 & 34.71 & \bf41.65 & 64.17 & 66.38 & 68.45 &\bf72.55 \\
\hline
\multirow{4}{*}{\textbf{Weather}} & Snow & 27.57 & 46.38 & 55.90 & 2.30 & 5.08 & 5.73 & 52.73 & 63.30 & 62.84 & \bf64.03 \\
& Rain & 27.71 & 46.50 & 56.08 & 13.51 & 20.39 & 24.97 & 58.40 & 65.35 & 66.13 & \bf69.09\\
& Fog & 24.49 & 41.64 & 43.78 & 12.83 & 27.89 & 32.76 & 53.19 & 53.67 & 54.10 & \bf61.80\\
& Sunlight & 23.71 & 40.28 & 54.20 & 22.77 & 34.66 & 41.68 & 57.70 & 55.14 & 64.42 & \bf70.88\\
\hline
\multirow{10}{*}{\textbf{Sensor}} & Density & 27.27 & 46.14 & 58.60 & - & - & - & 63.72 & 65.77 & 67.79 & \bf72.21\\
& Cutout & 24.14 & 40.95 & 56.28  & - & - & - & 62.25 & 63.66 & 66.18 & \bf70.92\\
& Crosstalk & 25.92 & 44.08 & 56.64 & - & - & - & 62.66 & 64.67 & 67.32 & \bf71.18 \\
& FOV Lost & 8.87 & 15.40 & 20.84  & - & - & - & 26.32 & 24.63 & 27.17 & \bf29.15\\
& Gaussian (L) & 19.41 & 39.16 & 45.79  & - & - & - & 58.94 & 55.10 & 60.64 & \bf68.84\\
& Uniform (L) & 25.60 & 45.00 & 56.12 & - & - & - & 63.21 & 64.72 & 66.81 & \bf71.82 \\
& Impulse (L) & 26.44 & 45.58 & 57.67 & - & - & - & 63.43 & 65.51 & 67.54 & \bf72.03\\
& Gaussian (C) & - & - & -  & 4.33 & 14.86 & 15.04 & 54.96 & 64.52 & 64.44 & \bf65.94\\
& Uniform (C) & - & - & -  & 8.48 & 21.49 & 23.00 & 57.61 & 65.26 & 65.81 & \bf68.30\\
& Impulse (C) & - & - & -  & 3.78 & 14.32 & 13.99 & 55.16 & 64.37 & 64.30 & \bf65.35\\
\hline
\multirow{3}{*}{\textbf{Motion}} & Compensation & 3.85 & 10.39 & 11.02 & - & - & - & 31.87 & 9.01 & 27.57 & \bf39.12\\
& Moving Obj. & 19.38 & 35.11 & 44.30 & 10.47 & 16.63 & 20.22 & 45.43 & 51.01 & \bf51.63 &50.08 \\
& Motion Blur & - & - & - & 9.64 & 11.06 & 19.79 & 55.99 & 64.39 & 64.74 & \bf67.03 \\
\hline
\multirow{8}{*}{\textbf{Object}} & Local Density & 26.70 & 45.42 & 57.55 & - & - & - & 63.60 & 65.65 & 67.42 &\bf71.63 \\
& Local Cutout & 17.97 & 32.16 & 48.36 & - & - & - & 61.85 & 63.33 & 63.41 & \bf67.79 \\
& Local Gaussian & 25.93 & 43.71 & 51.13 & - & - & - & 62.94 & 63.76 & 64.34 & \bf70.58\\
& Local Uniform & 27.69 & 46.87 & 57.87  & - & - & - & 64.09 & 66.20 & 67.58 & \bf72.29\\
& Local Impulse & 27.67 & 46.88 & 58.49 & - & - & - & 64.02 & 66.29 & 67.91 & \bf72.31\\
& Shear & 26.34 & 43.28 & 49.57  & 16.66 & 17.46 & 24.71 & 55.42 & \bf62.32 & 60.72 & 58.78\\
& Scale & 27.29 & 45.98 & 51.13  & 6.57 & 12.02 & 17.64 & 56.79 & 64.13 & 64.57 & \bf66.07\\
& Rotation & 27.80 & 46.93 & 54.68 & 16.84 & 27.28 & 33.97 & 59.64 & 63.36 & 65.13 & \bf69.49\\
\hline
\multirow{2}{*}{\textbf{Alignment}} & Spatial & - & - & - & - & - & - & 63.77 & 66.22 & 68.39 & \bf71.18\\
& Temporal& - & - & - & - & - & - & 51.43 & 43.65 & 49.02 & \bf56.64\\
\hline
\rowcolor{Gray}\multicolumn{2}{c|}{\textbf{Average} ($\mathrm{mAP_{cor}}$)} & 23.42 & 40.37 & \bf49.81 & 10.68 & 18.60 & \bf22.79 & 56.99 & 58.73 & 61.03&\bf64.98 \\
\hline
\end{tabular}
}
\end{center}
\end{table*}

We then analyze the performance under different corruptions, as shown in Table \ref{tab:10}. 
We conduct our experiments in a widely used benchmark with common corruptions
in autonomous driving, i.e., nuScenes-C \cite{dong2023benchmarking}.
There totally has 27 types of common corruptions for both LiDAR and camera inputs in real-world scenarios, which could be divided into five classifications, including, weather, sensor, motion, object, and alignment. It could be observed that our proposed CoreNet achieves the best performance by achieving 64.98\% mAP$_{cor}$. Particularly, in some extremely hard cases, our method shows great robustness. For example, CoreNet improves the mAP$_{cor}$ by 11.55 \% compared to BEVFusion \cite{liu2023bevfusion} under motion compensation.

\section{Conclusion}

In this paper, we proposed a novel model termed \textbf{Co}nflict \textbf{Re}solution \textbf{Net}work (CoreNet) for LiDAR-camera 3D object detection. We first conduct in-depth thinking on the task and summarize two conflicts, i.e., point-pixel misalignment and sub-task suppression that are overlooked by previous methods. Point-pixel misalignment ignores the occlusion and opacity relationship of objects and projects single-pixel features incorrectly, resulting in depth ambiguity and information interference. Sub-task suppression can lead to mutual disturbance between classification prediction and bounding box regression, resulting in sub-optimal results. To solve these issues, we introduce two novel modules: dual-stream transformation and task-specific predictor. Specifically, dual-stream transformation uses a two-branch structure to approximately scatter a pixel feature into a single point of the same ray, while task-specific predictor explicitly and controllably creates task-specific features and combines them with general features to develop task-specific queries. Extensive experiments are carried out on the nuScenes dataset to prove the superiority of our method, the effectiveness of each module, and the robustness of the method. We believe that CoreNet can provide useful inspiration for future methods.

\textbf{Limitations and further work.} Although our method achieves good performance, there is still room for improvement in some extremely difficult conditions such as detecting objects at night or at a far distance. In future work, we will introduce relevant modules and data augmentation methods to address these issues.  

% \section*{CRediT authorship contribution statement} 

% \textbf{Yiheng Li:} proposed the novel framework of solving 3D object detection in autonomous driving, conducted experiments and wrote the manuscript.
% \textbf{Yang Yang:} discussed and optimized the proposed framework to make the method achieve the SOTA results; wrote the introduction and partial method of the paper. 
% \textbf{Zhen Lei:} participated in the discussion of the contributions, made a suggestion about the organization structure of the paper.

\section*{Data availability}
    The code and data used in this study are available at \url{https://github.com/liyih/CoreNet}.

\section*{Declaration of competing interest}
    I have nothing to declare.

\section*{Acknowledgment}
This work was supported in part by the Chinese National Natural Science Foundation Project 62206276, 62276254, U23B2054, and the InnoHK program.
    
\appendix
% \section{Example Appendix Section}


{
\def\urlprefix{}
\bibliographystyle{elsarticle-num} 
\biboptions{sort&compress}
\bibliography{ref}
}

\end{document}